\documentclass{article}
\usepackage{spconf,amsmath,graphicx}
\usepackage{subcaption}
\graphicspath{{./images/}}
\usepackage{color}
\usepackage{multirow}
\usepackage{pifont}% http://ctan.org/pkg/pifont
\usepackage{lipsum}
\usepackage{adjustbox}

\let\OLDthebibliography\thebibliography
\renewcommand\thebibliography[1]{
  \OLDthebibliography{#1}
  \setlength{\parskip}{0pt}
  \setlength{\itemsep}{5pt}
}

\newcommand{\etal}{\textit{et al}.}
\newcommand{\bl}[1]{\textcolor{blue}{#1}}
\newcommand{\red}[1]{\textcolor{red}{#1}}
\newcommand{\cmark}{\ding{51}}%
\newcommand{\xmark}{\ding{55}}%

\title{raising The Limit Of Image Rescaling Using Auxiliary Encoding}
\name{Chenzhong Yin$^{1,*}$ \qquad Zhihong Pan$^{2, }$\sthanks{Both authors contributed equally to this work when Chenzhong Yin interned at Baidu} \qquad Xin Zhou$^{2}$ \qquad Le Kang$^{2}$ \qquad Paul Bogdan$^{1}$}

\address{$^{1}$University of Southern California, Los Angeles, CA, USA \\
$^{2}$Baidu Research (USA), Sunnyvale, CA, 94089, USA}
\begin{document}
%\ninept
%
\maketitle
\begin{abstract}
Normalizing flow models using invertible neural networks (INN) have been widely investigated for successful generative image super-resolution (SR) by learning
the transformation between the normal distribution of latent variable $z$ and the conditional distribution of high-resolution
(HR) images gave a low-resolution (LR) input.  Recently, image rescaling models like IRN utilize the bidirectional nature of INN to push the performance limit of image upscaling
by optimizing the downscaling and upscaling steps jointly.  While the random sampling of latent variable $z$ is useful in generating diverse
photo-realistic images, it is not desirable for image rescaling when accurate restoration of the HR image is more important.
Hence, in places of random sampling of $z$, we propose auxiliary encoding modules to further push the limit of image rescaling performance.
Two options to store the encoded latent variables in downscaled LR images, both readily supported in existing image file format,
are proposed. 
One is saved as the alpha-channel, the other is saved as meta-data in the image header, and the corresponding modules are denoted as suffixes -A and -M respectively.
Optimal network architectural changes are investigated for both options to demonstrate their effectiveness
in raising the rescaling performance limit on different baseline models including IRN and DLV-IRN.
%It is also validated that these enhancement is applicable to other INN based rescaling models like DLV-LRN.

\end{abstract}


\vspace{-2mm}
\section{Introduction}
\vspace{-1mm}
Currently, ultra-high resolution (HR) images are often needed to be reduced from their original resolutions to lower ones due to various limitations like display or transmission.
Once resized, there could be subsequent needs of scaling them up so it is useful to restore more high-frequency details~\cite{xing_arxiv_2022}.
While deep learning super-resolution (SR) models~\cite{dong2014learning, zhang_eccv_2018, yin2023anatomically} are powerful tools to reconstruct HR images from low-resolution (LR) inputs, they are often limited to pre-defined image downscaling methods. Additionally, due to memory and speed constraints, HR images or videos are also commonly resized to lower resolution for downstream computer vision tasks
like image classification and video understanding. Similarly, they rely on conventional resizing methods which are subject to information loss and have negative impact on downstream tasks~\cite{tian_cvpr_2021}.
Hence, learned image downscaling techniques with minimum loss in high-frequency information are quite indispensable for both scenarios.
Lastly, it is known that SR models optimized for upscaling only are subject to model stability issues when multiple downscaling-to-upscaling cycles are applied~\cite{pan_cvpr_2022} so it further
validates the necessity of learning downscaling and upscaling jointly.

 \begin{figure*}[ht!]
 \begin{center}
     \includegraphics[width=\linewidth]{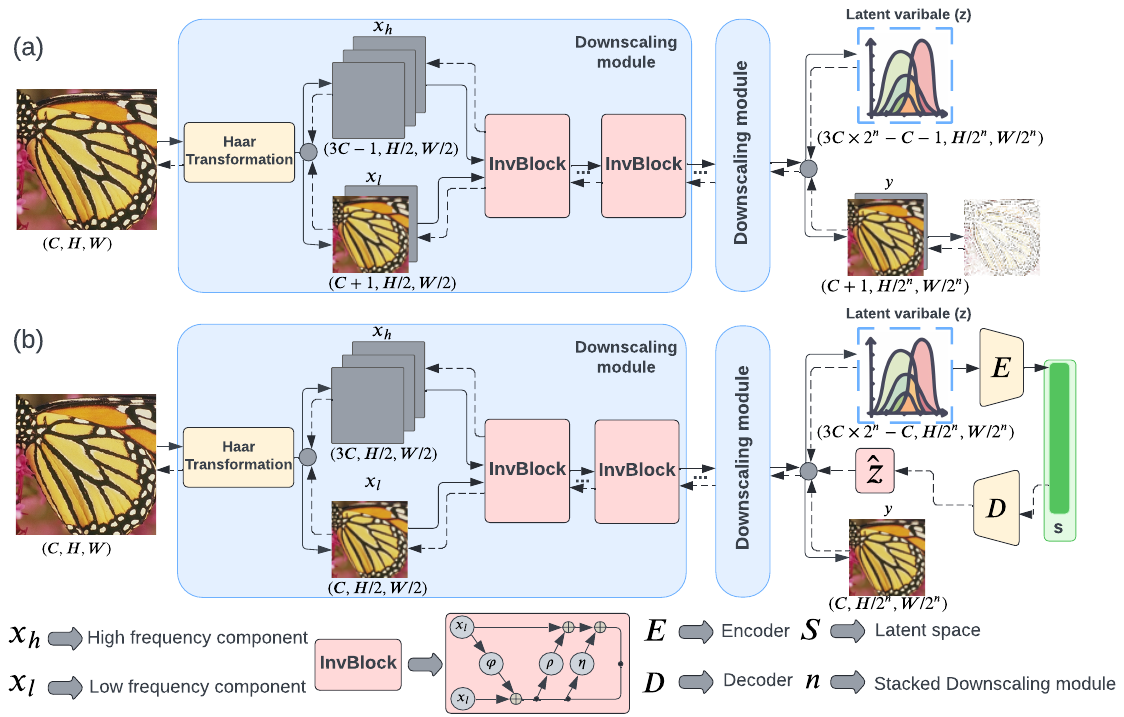}
 \end{center}
 \vspace{-10pt}
 \caption{Illustration of invertible image rescaling network architecture: (a) RGBA approach and (b) metadata approach.}
 %Invertible
%neural network blocks (InvBlocks) are placed after a Haar Transformation } 
 \vspace{-10pt}
 \label{fig:net}
 \end{figure*}
 
%  \begin{figure}[h!]
 %
%  \begin{center}
%      \includegraphics[width=0.75\linewidth]{images/AE.pdf}
%  \end{center}
%  %
%  \vspace{-10pt}
%  \caption{Autoencoder architecture used in Fig.~\ref{fig:net} (b)}
%  \vspace{-10pt}
%  \label{fig:ae}
%  %
%  \end{figure}

To overcome these challenges and utilize the relationship between upscaling and downscaling steps, recent works designed the encoder-decoder framework to unite these two independent tasks together.
Kim \etal~\cite{kim2018task} utilized autoencoder (AE) architecture, where the encoder is the downscaling network and the decoder is the upscaling network, to find the optimal LR result that maximizes the restoration performance of the HR image. Sun \etal~\cite{sun2020learned} designed a learned content adaptive image downscaling model in which an SR model is trained simultaneously to best recover the HR images. 
Later on, Li \etal~\cite{li2018learning} proposed a learning approach for image compact-resolution using a convolutional neural network (CNN-CR) where the image SR problem is formulated to jointly minimize the reconstruction loss and the regularization loss. Although the above models can efficiently improve the quality of HR images recovered from corresponding LR images, these works only optimize downscaling and SR separately, while ignoring the potential mutual intension between downscaling and inverse upscaling. 

More recently, a jointly optimized rescaling model was proposed by Xiao~\etal~\cite{xiao2020invertible}
to achieve significantly improved performance.
%They designed an Invertible Rescaling Net (IRN) to capture the lost information in the form of its distribution and embedded it into the model’s parameters to mitigate the ill-posedness.
An Invertible Rescaling Net (IRN) was designed to model the reciprocal nature of the downscaling and upscaling processes.
For downscaling, IRN was trained to convert HR input to visually-pleasing LR output and a latent variable $z$.
As $z$ is trained to follow an input-agnostic Gaussian distribution, the HR image can be accurately reconstructed
during the inverse up-scaling procedure although $z$ is randomly sampled from a normal distribution. Nevertheless,
the model's performance can be further improved if the high-frequency information remaining in $z$ is efficiently stored.  

To resolve the above difficulties and take full potential of the IRN, here we propose two approaches, namely the IRN-meta (IRN-M) and IRN-alpha (IRN-A), respectively, to efficiently compress the high frequency information stored in $z$, which can be used to recover $z$
and help restore HR image consequently during the inverse up-scaling.
%IRN-A saves the LR output in RGBA format where the alpha-channel stores the compressed $z$, while IRN-M uses AE to compress $z$ as a latent variable to store as meta-data in LR output.
For IRN-A, we train the model to extract a fourth LR channel in addition to the LR RGB channels.
It represents essential high frequency information which was lost in the IRN baseline due to random sampling of $z$,
and is saved as the alpha-channel of saved LR output.
For IRN-M approach, an AE module is trained to compress $z$ as a compact latent variable, which can be saved
as metadata of the LR output. In the inverse upscaling process, $z$ is restored from the latent space by utilizing the well-trained decoder.
Both modules are also successfully applied to the state-of-the-art (SOTA) rescaling model DLV-IRN~\cite{zhang2022enhancing}.
In summary, the main contribution of this paper is that we are the first to compress the high-frequency information in $z$,
which is not fully utilized in current invertible image rescaling models, to improve the restored HR image quality in upscaling progress. 

\vspace{-2mm}
\section{Proposed Method}
\label{sec:method}
%\subsection{Problem Specification}
\vspace{-2mm}
\subsection{IRN-A}
\vspace{-1mm}
Fig.~\ref{fig:net} (a) shows the IRN-A network architectures, where the invertible neural network blocks (InvBlocks) are referenced from previous work IRN~\cite{xiao2020invertible}. In the new model, the input HR image is resized via Haar transformation before
splitting to a lower branch $x_l$ and a higher branch $x_h$.
More specifically, Haar transformation converts the input HR image $(C, H, W)$ into a matrix of shape $(4C, H/2, W/2)$, where $C$, $H$, and $W$ represent image color channels, height and width respectively. The first $C$ channels represent low-frequency components of the input image in general
and the remaining $3C$ channels represent the high-frequency information on vertical, horizontal and diagonal directions respectively.
Different from the IRN baseline, which uses only the $C$ low-frequency channels in the lower branch,
we add $1$ additional channel, denoted as alpha-channel for convenience as it would be stored as the alpha-channel in RGBA format,
in the lower branch $x_l$ to store the compressed high-frequency information.
After the first Haar transformation, the alpha-channel is initialized with the average value across all $3C$ high-frequency channels,
and only $3C-1$ channels are included in $x_h$ as the first channel is removed to make the total number of channels remain constant. 

\vspace{-1.mm}
After channel splitting, $x_l$ and $x_h$ are fed into cascaded InvBlocks and transformed to an LR RGBA image $y$ and an auxiliary
latent variable $z$.
First three channels of $y$ consist of the visual RGB channels and the fourth channel contains the compressed high-frequency
components transformed along the InvBlocks. The alpha-channel was normalized via a $sigmoid$ function, $S(\alpha)=\frac{1}{1+e^{-\alpha}}$, to help quantization
of the alpha-channel and maintain training stability.
%\begin{equation}
%\vspace{-3pt}
%    S(\alpha)=\frac{1}{1+e^{-\alpha}}
%\vspace{0pt}
%\label{eq:loss}
%\end{equation}

For the inverse upscaling process,
the model needs to recover $z$, denoted as $\hat{z}$ as it is not stored. In previous work, $\hat{z}$ was randomly drawn from normal Gaussian distribution.
While this helps creating diverse samples in generative models, it is not optimal for tasks like image rescaling which aims to restore one HR image instead
of diverse variations.
% The main limitation of this approach is that the randomness may cause the training process to become unstable and influence the model performance.
Therefore, we set $\hat{z}$ as $0$, the mean value of the normal distribution, for the inverse up-scaling process.
This technique was also validated in previous works like FGRN~\cite{li2021approaching} and DLV-IRN~\cite{zhang2022enhancing}.
Of note, at the end of inverse process, the deleted high frequency channel needs to be recovered as
\begin{equation}
\vspace{-4pt}
    x_{m}=3C\times x_{\alpha}-\sum\nolimits_{i=1}^{3C-1} {x^i_h}
\vspace{0pt}
\label{eq:recover}
\end{equation}
where $x_m$ represents the channel removed from $x_h$ and $x_{\alpha}$ represents the alpha-channel in $x_l$.

\vspace{-2mm}
\subsection{IRN-M}
\vspace{-1mm}
Besides storing the compressed high-frequency information in a separate alpha-channel, we also propose an alternative space-saving approach to store the extracted information as metadata of the image file. Image metadata is text information pertaining to an image file that is embedded into the image file or contained in a separate file in a digital asset management system.  Metadata is readily supported by existing image format so this proposed method could be easily integrated with current solutions.  

The network architecture of our metadata approach is shown in Fig.~\ref{fig:net} (b). Here $x_l$ and $x_h$, same as the IRN baseline,
are split from Haar transformed $4C$ channels to $C$ and $3C$ channels respectively.
Unlike the RGBA approach, the metadata method uses an encoder at the end to compress the $z$ and save the latent vector $S$ as metadata,
rather than saving as the alpha-channel of the output. $S$ will be decompressed by the decoder for the inverse upscaling step.
In our AE architecture, the encoder compacts the number of $z$ channels from $3C\times n^2 -C$ to 4 via 2D convolution layers and
compresses the $z$'s height and width from $(H/2^{n}, W/2^{n})$ to $(H/2^{n+2}, W/2^{n+2})$ by using max-pooling layers.
Here $n$ is 1 or 2 depending on the scale factor of $2\times$ or $4\times$.
Of note, the AE was pre-trained with MSE loss before being embedded into the model structure.

After placing the well-trained AE in the IRN architecture, the entire structure was trained to minimize the following mixture loss function:
\begin{equation}
\vspace{-4pt}
     L = \lambda_1 L_{r}+\lambda_2 L_{g}+\lambda_3 L_{d}+\lambda_4 L_{mse}
\vspace{-0pt}
\label{eq:loss}
\end{equation}
where $L_{r}$ is the $L1$ loss for reconstructing HR image; $L_g$ is the $L2$ loss for the generated LR image; $L_d$ is the distribution matching loss; and $L_{mse}$ is the MSE loss between the input of the encoder and the output of the decoder. 
\begin{table}[h!]
\small
 \caption{Comparison of 4$\times$ upscaling results using different IRN-A hyperparameters and settings. The best results are highlighted in \textcolor{red}{red}.}
\vspace{-1mm}
\begin{adjustbox}{width=\columnwidth,center}
%\begin{center}
\begin{tabular}{c|c|c|c|c}
\hline \hline
\multirow{2}{*}{IRN-A}                     & \multirow{2}{*}{$\alpha_{avg}$}& BSD100       & Urban100     & DIV2K        \\
%\hline
\cline{3-5}& & PSNR/SSIM$\uparrow$ & PSNR/SSIM$\uparrow$ & PSNR/SSIM$\uparrow$ \\
\hline
{Post-split}     & \xmark               & 32.66 / 0.9083 & 32.50 / 0.9328 & 36.19 / 0.9464 \\
                            \hline
\multirow{2}{*}{Pre-split}
                               & \xmark               & 33.02 / 0.9132 & 32.17 / 0.9186 & 36.60 / 0.9495 \\
                               & \cmark              & \red{33.12} / \red{0.9150} & \red{33.10} / \red{0.9384} & \red{36.67} / \red{0.9504}\\
\hline\hline

\end{tabular}
%\end{center}
\end{adjustbox}
\label{tab:1}
\end{table}

% \usepackage{multirow}
\begin{table}[h!]
\caption{Comparison of 4$\times$ upscaling results using different IRN-M hyperparameters and settings. The best results are highlighted in \textcolor{red}{red}.}
\vspace{-1mm}
\begin{adjustbox}{width=\columnwidth,center}
\begin{tabular}{c|c|c|c|c|c}
\hline \hline
\multirow{2}{*}{IRN-M}                 & \multirow{2}{*}{$AE_{p}$} & \multirow{2}{*}{$AE_{f}$} & BSD100       & Urban100     & DIV2K        \\
\cline{4-6}& & & PSNR/SSIM$\uparrow$ & PSNR/SSIM$\uparrow$ & PSNR/SSIM$\uparrow$ \\
\hline
\multirow{3}{*}{2layers} &  \xmark           &  \xmark       & 31.41 / 0.8771 & 30.79 / 0.9074 & 34.79 / 0.9283 \\
                         & \cmark           & \cmark       & 31.58 / 0.8793 & 31.30 / 0.9123 & 35.06 / 0.9306 \\
                         & \cmark           &  \xmark       & 31.65 / 0.8804 & 31.34 / 0.9154 & 35.09 / 0.9306 \\
                         \hline
\multirow{2}{*}{4layers}                   &  \xmark           &  \xmark       & 28.15 / 0.7765 & 25.82 / 0.7989 & 30.72 / 0.8591 \\
                         %& \cmark           & \cmark       & 31.29 / 0.8720 & 30.66 / 0.9031 & 34.60 / 0.9495 \\
                         & \cmark           &  \xmark       & \red{31.69} / \red{0.8812} & \red{31.44} / \red{0.9143} & \red{35.15} / \red{0.9314}\\
                         \hline \hline
                        
\end{tabular}
\end{adjustbox} 
\label{tab:2}
\end{table}

\vspace{-2mm}
\section{Experiments}
\vspace{-2mm}
Following the same training strategy and hyperparameters in IRN baseline, our models were trained on the DIV2K~\cite{bell2019blind} dataset, which includes $800$ HR training images. IRN-M and IRN-A were trained with 500,000 and 250,000 iterations respectively.
Both models were evaluated across five benchmark datasets: Set5~\cite{bevilacqua_bmvc_2012}, Set14~\cite{zeyde_iccs_2010}, BSD100~\cite{martin_iccv_2001}, Urban100~\cite{huang_cvpr_2015} and the validation set of DIV2K. The upscaled images quality across different models were assessed via the peak noise-signal ratio (PSNR) and SSIM on the Y channel of the YCbCr color space.
Following previous works~\cite{li2021approaching, zhang2022enhancing}, as it is not beneficial to add
randomness in restoring HR images, we set $\hat{z}$ as $0$ during  the inverse up-scaling process
for both training and validation steps in all experiments.

\begin{figure*}[ht!]
\captionsetup[subfigure]{font=footnotesize, labelformat=empty}
\begin{center}
  \begin{subfigure}[b]{0.165\textwidth}
    \centering
      \includegraphics[width=\textwidth, interpolate=false]{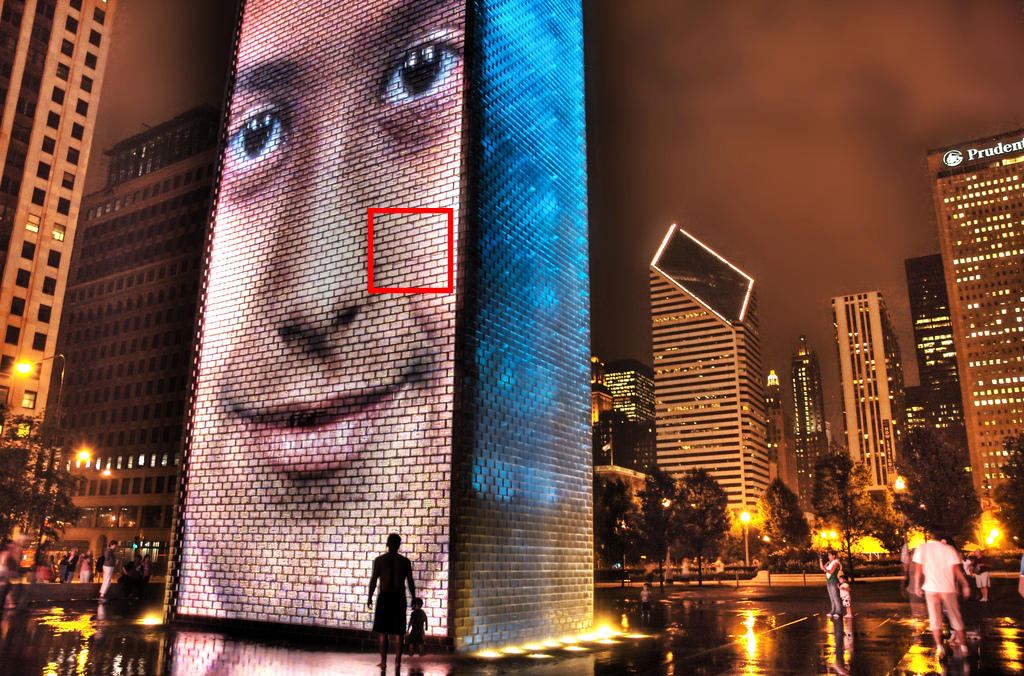}
  \end{subfigure} \hspace*{-0.45em}
  \begin{subfigure}[b]{0.11\textwidth}
    \centering
      \includegraphics[width=\textwidth, interpolate=false]{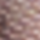}
  \end{subfigure} \hspace*{-0.45em}
  \begin{subfigure}[b]{0.11\textwidth}
    \centering
      \includegraphics[width=\textwidth, interpolate=false]{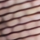}
  \end{subfigure} \hspace*{-0.45em}
  \begin{subfigure}[b]{0.11\textwidth}
    \centering
      \includegraphics[width=\textwidth, interpolate=false]{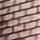}
  \end{subfigure} \hspace*{-0.45em}
  \begin{subfigure}[b]{0.11\textwidth}
    \centering
      \includegraphics[width=\textwidth, interpolate=false]{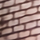}
  \end{subfigure} \hspace*{-0.45em}
  \begin{subfigure}[b]{0.11\textwidth}
    \centering
      \includegraphics[width=\textwidth, interpolate=false]{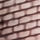}
  \end{subfigure} \hspace*{-0.45em}
  \begin{subfigure}[b]{0.11\textwidth}
    \centering
      \includegraphics[width=\textwidth, interpolate=false]{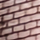}
  \end{subfigure} \hspace*{-0.45em}
  \begin{subfigure}[b]{0.11\textwidth}
    \centering
      \includegraphics[width=\textwidth, interpolate=false]{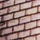}
  \end{subfigure}
%  \begin{subfigure}[b]{0.165\textwidth}
%    \centering
%      \includegraphics[width=\textwidth, interpolate=false]{images/44_GT.png}
%  \end{subfigure} \hspace*{-0.45em}
%  \begin{subfigure}[b]{0.11\textwidth}
%    \centering
%      \includegraphics[width=\textwidth, interpolate=false]{images/car_44.png}
%  \end{subfigure} \hspace*{-0.45em}
%  \begin{subfigure}[b]{0.11\textwidth}
%    \centering
%      \includegraphics[width=\textwidth, interpolate=false]{images/irn_44.png}
%  \end{subfigure} \hspace*{-0.45em}
%  \begin{subfigure}[b]{0.11\textwidth}
%    \centering
%      \includegraphics[width=\textwidth, interpolate=false]{images/a_hcflow_44.png}
%  \end{subfigure} \hspace*{-0.45em}
%  \begin{subfigure}[b]{0.11\textwidth}
%    \centering
%      \includegraphics[width=\textwidth, interpolate=false]{images/dlvrin_44.png}
%  \end{subfigure} \hspace*{-0.45em}
%  \begin{subfigure}[b]{0.11\textwidth}
%    \centering
%      \includegraphics[width=\textwidth, interpolate=false]{images/irnm_44.png}
%  \end{subfigure} \hspace*{-0.45em}
%  \begin{subfigure}[b]{0.11\textwidth}
%    \centering
%      \includegraphics[width=\textwidth, interpolate=false]{images/irna_44.png}
%  \end{subfigure} \hspace*{-0.45em}
%  \begin{subfigure}[b]{0.11\textwidth}
%    \centering
%      \includegraphics[width=\textwidth, interpolate=false]{images/gt_44.png}
%  \end{subfigure}
  \begin{subfigure}[b]{0.165\textwidth}
    \centering
      \includegraphics[width=\textwidth, interpolate=false]{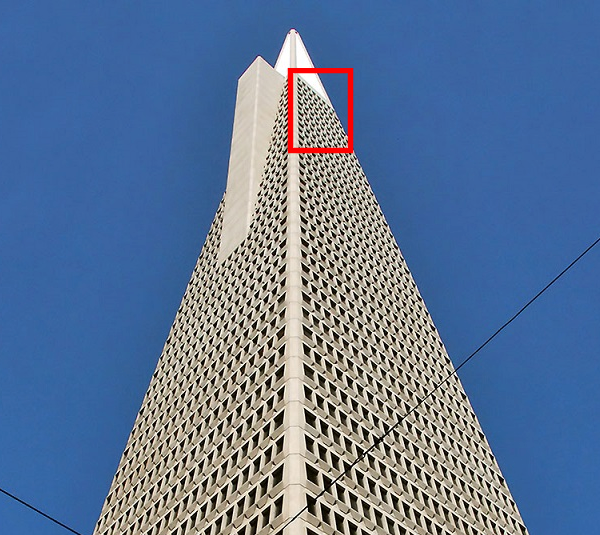}
      \caption{HR Image}
  \end{subfigure} \hspace*{-0.45em}
  \begin{subfigure}[b]{0.11\textwidth}
    \centering
      \includegraphics[width=\textwidth, interpolate=false]{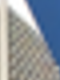}
      \caption{Bicubic}
  \end{subfigure} \hspace*{-0.45em}
  \begin{subfigure}[b]{0.11\textwidth}
    \centering
      \includegraphics[width=\textwidth, interpolate=false]{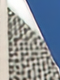}
      \caption{CAR~\cite{sun2020learned}}
  \end{subfigure} \hspace*{-0.45em}
  \begin{subfigure}[b]{0.11\textwidth}
    \centering
      \includegraphics[width=\textwidth, interpolate=false]{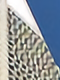}
      \caption{IRN~\cite{xiao2020invertible}}
  \end{subfigure} \hspace*{-0.45em}
  \begin{subfigure}[b]{0.11\textwidth}
    \centering
      \includegraphics[width=\textwidth, interpolate=false]{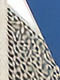}
      \caption{DLV-IRN~\cite{zhang2022enhancing}}
  \end{subfigure} \hspace*{-0.45em}
  \begin{subfigure}[b]{0.11\textwidth}
    \centering
      \includegraphics[width=\textwidth, interpolate=false]{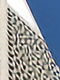}
      \caption{DLV-IRN-M}
  \end{subfigure} \hspace*{-0.45em}
  \begin{subfigure}[b]{0.11\textwidth}
    \centering
      \includegraphics[width=\textwidth, interpolate=false]{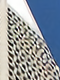}
      \caption{DLV-IRN-A}
  \end{subfigure} \hspace*{-0.45em}
  \begin{subfigure}[b]{0.11\textwidth}
    \centering
      \includegraphics[width=\textwidth, interpolate=false]{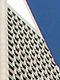}
      \caption{GT}
  \end{subfigure}
 \end{center}
   \vspace{-10pt}
    \caption{Visual examples from Urban100 test set (Best viewed in online version with zoom-in).}
   \vspace{-5pt}
\label{fig:x4}
\end{figure*}

\begin{table*}[h!]
\small
\setlength{\tabcolsep}{2.5pt}
   \vspace{2pt}
    \caption{Quantitative results of upscaled $\times4$ images of 5 datasets across different bidirectional rescaling approaches. The best two results highlighted in \red{red} and \bl{blue} respectively.}
    \vspace{-10pt}
\begin{center}
\begin{tabular}{l|c|c|c|c|c|c}
\hline \hline
%\multirow{2}{*}{Method} & \multicolumn{2}{c|}{Set5~\cite{bevilacqua_bmvc_2012}} & \multicolumn{2}{c|}{Set14~\cite{zeyde_iccs_2010}} & \multicolumn{2}{c|}{BSD100~\cite{martin_iccv_2001}} & \multicolumn{2}{c|}{Urban100~\cite{huang_cvpr_2015}} & \multicolumn{2}{c}{DIV2K~\cite{agustsson_ntire_2017}}\\
\multirow{2}{*}{Method}&\multirow{2}{*}{Scale} &{Set5~\cite{bevilacqua_bmvc_2012}} &{Set14~\cite{zeyde_iccs_2010}} &{BSD100~\cite{martin_iccv_2001}} &{Urban100~\cite{huang_cvpr_2015}} &{DIV2K~\cite{agustsson_ntire_2017}}\\
\cline{3-7} & & PSNR/SSIM$\uparrow$ & PSNR/SSIM$\uparrow$ & PSNR/SSIM$\uparrow$ & PSNR/SSIM$\uparrow$ & PSNR/SSIM$\uparrow$ \\
\hline
%Bicubic \& Bicubic~\cite{zhang_eccv_2018} & 2&33.66 / 0.9299 & 30.24 / 0.8688 & 29.56 / 0.8431 & 26.88 / 0.8403 & 31.01 / 0.9393 \\
CAR~\cite{sun2020learned}& 2& 
38.94 / 0.9658& 35.61 / 0.9404& 33.83 / 0.9262& 35.24 / 0.9572& 38.26 / 0.9599\\
%\hline
IRN~\cite{xiao2020invertible}& 2    & 43.99 / 0.9871 & 40.79 / 0.9778 &41.32 / 0.9876&39.92 / 0.9865&44.32 / 0.9908 \\
FGRN~\cite{li2021approaching}&2&44.15 / 0.9902& 42.28 / 0.9840& 41.87 / 0.9887& 41.71 /  0.9904&45.08 / 0.9917\\
DLV-IRN~\cite{zhang2022enhancing}& 2& 45.42 / 0.9910& 42.16 / 0.9839& 42.91 / 0.9916&41.29 / 0.9904&45.58 / 0.9934\\
\textbf{DLV-IRN-M}&2 & \bl{45.83} / \bl{0.9916} & \bl{42.47} / \bl{0.9850} & \bl{43.38} / \bl{0.9925} & \bl{41.77} / 
\bl{0.9911} & \bl{45.91} / \bl{0.9939}\\
\textbf{DLV-IRN-A}&2& \red{47.81} / \red{0.9937} & \red{44.96} / \red{0.9884} & \red{47.15} / \red{0.9967} & \red{45.07} / \red{0.9953} & \red{48.94} / \red{0.9968} \\
\hline 
%Bicubic \& Bicubic~\cite{zhang_eccv_2018} & 4&
%28.42 / 0.8104 & 26.00 / 0.7027& 25.96 / 0.6675 & 23.14 / 0.6577 &26.66 / 0.8521\\
CAR~\cite{sun2020learned}& 4& 
33.88 / 0.9174& 30.31 / 0.8382& 29.15 / 0.8001& 29.28 / 0.8711& 32.82 / 0.8837\\
IRN~\cite{xiao2020invertible}&4&
36.19 / 0.9451& 32.67 / 0.9015& 31.64 / 0.8826& 31.41 / 0.9157& 35.07 / 0.9318\\
HCFlow~\cite{liang2021hierarchical}&4&
36.29 / 0.9468& 33.02 / 0.9065& 31.74 / 0.8864& 31.62 / 0.9206& 35.23 / 0.9346\\
FGRN~\cite{li2021approaching}&4&
\bl{36.97} / \bl{0.9505}& \bl{33.77} / \bl{0.9168}& 31.83 / 0.8907& 31.91 / 0.9253& 35.15 / 0.9322\\
{DLV-IRN}~\cite{zhang2022enhancing}&4&
36.62 / 0.9484& 33.26 / 0.9093& 32.05 / 0.8893& 32.26/ 0.9253& 35.55/ 0.9363\\
\textbf{DLV-IRN-M}&4&
36.67 / 0.9490 & 33.33 / 0.9105 & \bl{32.12} / \bl{0.8909} & \bl{32.33} / \bl{0.9264} & \bl{35.63} / \bl{0.9373}\\
\textbf{DLV-IRN-A}& 4 &
\red{37.56} / \red{0.9566} & \red{34.12} / \red{0.9246} & \red{33.12} / \red{0.9150} & \red{33.10} / \red{0.9384} & \red{36.67} / \red{0.9504} \\ 

\hline\hline
\end{tabular}
\end{center}
 
\label{tab:res}
\vspace{-12pt}
\end{table*}

\vspace{-2mm}
\subsection{Ablation study}
\vspace{-1mm}
As the transformed
alpha-channel is the key innovation for improved performance for IRN-A,
the pre-splitting and initial settings of the alpha-channel
before the forward transformation process are very important.
%Since it is not beneficial for adding randomness in restoring HR images, choosing fixed $z$ with a constant zero can stabilize the training and validation steps.
For better analysis of their effects, Table~\ref{tab:1} shows an ablation study that compares the results for different settings of the alpha-channel, where
``post-split" and ``pre-split" refer to splitting the alpha-channel after the downscaling module or before the InvBlock respectively, and
$\alpha_{avg}$ represents presetting the average value of high-frequency information in the pre-split alpha-channel.
From Table~\ref{tab:1}, we notice that using the $\alpha_{avg}$ with pre-split architecture performs best across all options. 

The IRN-M model constructs the HR image by decoding the latent space $s$ saved in the metadata file. Table~\ref{tab:2} shows another ablation study for determining the optimal AE structure, where $AE_{p}$ represents that AE, before training as part of IRN-M, is pre-trained using MSE loss with standalone random $z$;
$AE_{f}$ represents fixing the AE during training the IRN-M; and ``2layers" and ``4layers" represent two and four convolutional layers used in AE respectively.
As shown in Table~\ref{tab:2}, using the IRN-M with pre-trained 4 layers AE and not fixing the AE during training has the best performance.
Of all three settings, pre-training of AE is the most critical factor in maximizing performance.
% \usepackage{multirow}

\vspace{-4mm}
\subsection{Image rescaling}
\vspace{-1mm}
The quantitative comparison results for HR image reconstruction are shown in Table~\ref{tab:res}. 
Rather than choosing SR models which only optimize upscaling steps, we consider SOTA bidirectional (jointly optimizing downscaling and upscaling steps) models for fair comparison~\cite{sun2020learned, xiao2020invertible, zhang2022enhancing, li2021approaching, liang2021hierarchical}. 
As shown in Table~\ref{tab:res}, DLV-IRN-A is efficient at storing high-frequency information in the alpha-channel and consequently outperforms its baseline DLV-IRN, as well as other models,
including HCFlow and IRN models, which randomly samples $\hat{z}$ for the upscaling step.
For DLV-IRN-M, while not as good as the -A variant, 
%when comparing with the second best FGRN,  which is an encoding-decoding based network without involving latent variable,
it still performs better  than all other models, only trailing behind FGRN for two small test sets at $4\times$.
%across all large test datasets (e.g., BSD100, Urban100 and DIV2K).
Hence we conclude that both -M and -A modules can improve the modeling of the high-frequency information and
help restore the HR image consequently.  Visual examples of the $4\times$ test in Fig~\ref{fig:x4} also validate the improved performance from our models.

\vspace{-8mm}
\section{Conclusions}
\vspace{-3mm}
To fully mine the potential of image rescaling models based on INN, two novel modules are proposed to store otherwise lost high-frequency information $z$.  The IRN-M model utilizes an autoencoder to compress
$z$ and save as metadata in native image format so it can be decoded to an approximate of $z$, while IRN-A adds an additional channel to store crucial high-frequency information, which can be quantized and stored
as the alpha-channel, in addition to the RGB channels, in existing RGBA format.  With carefully designed autoencoder and alpha-channel pre-split,
it is shown that both modules can improve the upscaling performance significantly comparing to the IRN baseline.
The proposed modules are also applicable to newer baseline models like DLV-IRN and DLV-IRN-A is by far the best, which further pushes the limit of image rescaling performance with a significant margin.

\bibliographystyle{IEEEbib}
\bibliography{mybib_2023}

\end{document}